\documentclass[letterpaper, 10 pt, conference]{ieeeconf}
\IEEEoverridecommandlockouts
\let\labelindent\relax

\usepackage[table,xcdraw]{xcolor} 
\usepackage{multicol}

\usepackage[caption=false,font=footnotesize,labelfont=sf,textfont=sf]{subfig}
\usepackage{graphicx}
\usepackage{standalone}
\usepackage{tikz}
\usetikzlibrary{shapes,backgrounds,shapes.geometric}
\usepackage{circuitikz}
\tikzset{font={\fontsize{9pt}{9}\selectfont}}
\usepackage{pgfplots}
\usepgfplotslibrary{fillbetween,colormaps,groupplots}
\pgfplotsset{compat=newest}

\usepackage{float}

\usepackage{setspace}

\usepackage{tabularx}
\usepackage{makecell}
\usepackage{booktabs}
\usepackage{multirow}
\newcolumntype{Y}{>{\raggedleft\arraybackslash}X}
\newcolumntype{Z}{>{\centering\arraybackslash}X}
\newcolumntype{L}{>{\hsize=0.61\hsize}X}
\newcolumntype{C}{>{\hsize=1\hsize}Z}
\newcolumntype{R}{>{\hsize=0.61\hsize}Y}
\newcolumntype{K}{>{\hsize=0.33333\hsize}X}
\newcolumntype{M}{>{\hsize=0.33333\hsize}Z}
\newcolumntype{N}{>{\hsize=0.33333\hsize}Y}

\usepackage{siunitx}
\usepackage{amsmath}
\usepackage{amsfonts}
\usepackage{scalerel}
\usepackage{amssymb}
\usepackage{bm}

\usepackage{algorithm}
\usepackage{algorithmicx}

\usepackage[utf8]{inputenc}
\usepackage{enumitem}

\date{\today}

\usepackage{pdfpages}
\usepackage[
	pdfauthor={Frederik Westerhout},
	pdfcreator={pdfLatex},
	pdfsubject={MSc Thesis Robotics},
    colorlinks={true},
    linkcolor={blue},
    citecolor={blue},
    urlcolor={cyan},
    hyperindex={true}
]{hyperref}
\makeatletter
\hypersetup{pdftitle=\@title}
\makeatother
\usepackage{mathtools}

\usepackage{tablefootnote}
\usepackage[caption=false, font=footnotesize]{subfig}

\newcommand\xlabel[2][]{\phantomsection\def\@currentlabelname{#1}\label{#2}}

\usepackage{cite}
\usepackage{jabbrv}

\usepackage{flushend}
\begin{document}

\title{Smooth-Trajectron++: Augmenting the Trajectron++ behaviour prediction model with smooth attention}

\makeatletter
\newcommand{\linebreakand}{%
  \end{@IEEEauthorhalign}
  \hfill\mbox{}\par
  \mbox{}\hfill\begin{@IEEEauthorhalign}
}
\makeatother

\author{%
    Frederik S.B. Westerhout\textsuperscript{1}
    \and
    Julian F. Schumann\textsuperscript{1}
    \and
    Arkady Zgonnikov\textsuperscript{1}
}

\setcounter{page}{1}
\maketitle
\footnotetext[1]{Department of Cognitive Robotics, Delft University of Technology, Delft, The Netherlands}

\thispagestyle{plain}
\pagestyle{plain}

\begin{abstract}
\setstretch{1.0} Understanding traffic participants' behaviour is crucial for predicting their future trajectories, aiding in developing safe and reliable planning systems for autonomous vehicles. Integrating cognitive processes and machine learning models has shown promise in other domains but is lacking in the trajectory forecasting of multiple traffic agents in large-scale autonomous driving datasets. This work investigates the state-of-the-art trajectory forecasting model Trajectron++ which we enhance by incorporating a smoothing term in its attention module. This attention mechanism mimics human attention inspired by cognitive science research indicating limits to attention switching. We evaluate the performance of the resulting Smooth-Trajectron++ model and compare it to the original model on various benchmarks, revealing the potential of incorporating insights from human cognition into trajectory prediction models.
\end{abstract}

\setcounter{footnote}{1}

\section{Introduction} 


In a world where the demand for intelligent vehicles is increasing rapidly~\cite{singh_mutreja}, the concern for the safety of passengers and other road users should grow with it~\cite{koopman2017autonomous}. According to the World Health Organization, approximately 1.3 million people die each year due to road traffic accidents, and this number is expected to increase if proper measures are not taken \cite{who-rti}. Therefore, it should be paramount for future autonomous vehicles to improve traffic safety.


One of the most critical factors for ensuring a safe environment around intelligent vehicles is accurately predicting the future movements of surrounding traffic participants. These predictions allow for a better assessment of the environment and anticipation of potentially dangerous situations at an early stage, lowering the risk of accidents. The accurate predictions of interactive behaviours are especially important, as those comprise the most challenging situations.



Numerous methods have been used to tackle the human behaviour prediction problem~\cite{camara_pedestrian_2021,bighashdel_survey_2019,rudenko_human_2019}, with examples ranging from reasoning-based methods to data-driven techniques. Over the last few years, data-driven approaches have shown great potential \cite{alahi_social_2016,mohamed_social-stgcnn_2020,giuliari_transformer_2021,salzmann_trajectron_2020,yuan2021agentformer,kalatian_context-aware_2022,rasouli_are_2017}, using machine learning algorithms to learn from large amounts of data to predict the trajectories of traffic participants. One of these data-driven models is \textit{Trajectron++} \cite{salzmann_trajectron_2020}, which stands out due to its public code availability, the general applicability and the results it achieved on multiple datasets (including \emph{nuScenes}~\cite{caesar_nuscenes_2020}, and \emph{highD}~\cite{krajewski2018highd}).

Other methods to predict future behaviour of traffic participants are based on theories from cognitive science. Instead of learning merely from data, the model is constructed to mimic human cognition. One class of such models based on the concept of evidence accumulation~\cite{gold2007neural} have proved useful specifically in predicting binary decisions in traffic interactions~\cite{zgonnikov_should_2022, schumann_using_2023}. However, these models are not yet applicable to trajectory forecasting in a more general setting. 
Another example of using insights from cognitive science for behaviour prediction in traffic is the use of a quantum-like Bayesian model, a mathematical framework that combines elements of quantum theory and Bayesian probability theory to describe decision-making and information processing in complex and uncertain environments \cite{moreira2017quantum}. It is used in~\cite{song_research_2022} to more accurately predict human street crossing behaviour, compared to the more data-driven model \emph{Social-LSTM}~\cite{alahi_social_2016}. Yet another insight from cognitive science suggests that the brain has a limited capacity for shifting attention rapidly between different tasks \cite{wolfe2000attention}. This is used in \cite{cao2022leveraging}, where the application of the smoothing term to the attention module of a machine-learning prediction model -- referred to as \emph{Smooth-Attention} -- which mimics human cognition, allows for better predictive performance.


Recent work demonstrated that integrating insights from cognitive science is a promising way of improving the performance of trajectory prediction models, but such cognitively inspired models need to be explored in a much more comprehensive way. Specifically, Cao et al.~\cite{cao2022leveraging} emphasize the need to combine smooth attention with more advanced interaction modelling network architectures. Here we aim to address this challenge by applying smooth attention to a state-of-the-art behaviour prediction model. Namely, we aim to improve upon the performance of \textit{Trajectron++} \textit{(T++)} \cite{salzmann_trajectron_2020} by leveraging the method of \emph{smooth attention} proposed in \cite{cao2022leveraging}. Applying a smoothness constraint on the attention module significantly reduces changes in attention, thereby ensuring that the module's output mimics the natural human cognitive processing. We name our approach of this combined model \textit{Smooth-Trajectron++}. 
We test this new model on the \emph{nuScenes}~\cite{caesar_nuscenes_2020} and \emph{highD}~\cite{krajewski2018highd} datasets.

\section{Methods} 

In this section, we provide an overview of the various elements of the \emph{Trajectron++} model (\emph{T++}) as well as the functioning of the \emph{Smooth-Attention} module.

\subsection{Trajectron++}
\label{edge:influence}
We use \textit{Trajectron++} \cite{salzmann_trajectron_2020} as the baseline model, for several reasons. Firstly, the model showed state-of-the-art performance on various public datasets \cite{salzmann_trajectron_2020} while including an attention module. Secondly, the authors have made the source code publicly available, including proper documentation regarding its application. This offers the opportunity to potentially reproduce the originally reported results while minimizing deviations from the original setup used by the authors. Lastly, the model has been tested on a large-scale public autonomous driving dataset, \emph{nuScenes} \cite{caesar_nuscenes_2020}. This provides evidence of the applicability of the model to real-world scenarios concerning interactions between multiple traffic participants.

\begin{figure}
    \centering
    \includegraphics[scale=0.55]{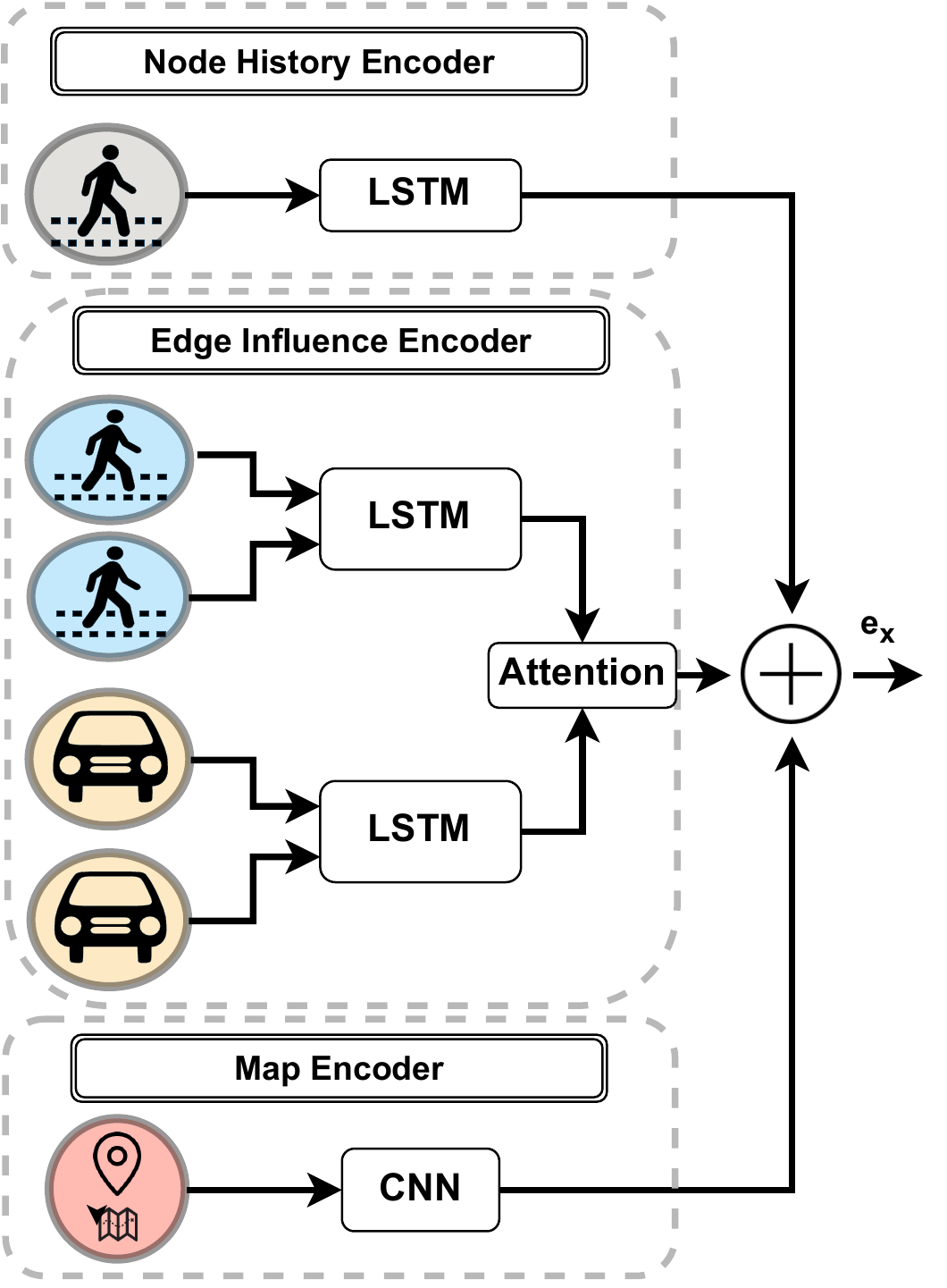}
    \caption{Encoder part of \emph{Trajectron++}~\cite{salzmann_trajectron_2020} that encodes various past input information into the representation vector \textbf{$e_x$}.}
     \vspace{-5mm}
    \label{fig:encoderT++}
\end{figure}

\emph{Trajectron++} is a graph-based conditional variational autoencoder model comprising an encoder and a decoder. The encoder uses various modules representing different influences on the trajectory forecast \autoref{fig:encoderT++}. First, the past location and speed of the chosen traffic agent $i$, for which predictions must be made, is fed into the Node History Encoder, whose main component is a long-short-term memory cell (LSTM). Second, the \textit{T++} model also uses road map data to make its predictions more feasible and realistic. 
The Map Encoder module takes relevant environmental information, fed to a convolutional neural network (CNN) and then added to the feature vector. Third, the Edge Influence Encoder models the influence of neighbouring agents on a given agent in the considered past time from ${t-T}$ to the current time ${t}$. To encode graph edges, the method follows a two-step process. Firstly, edge information is collected from neighbouring agents belonging to the same semantic class, such as pedestrian-pedestrian and car-car semantic classes. Summation is used for feature aggregation inside these classes to handle varying numbers of neighbouring nodes while preserving count information, following~\cite{jain2016structural}. The encodings of all the connections between the modelled agent and its neighbours are combined to create an "influence" representation vector, which represents the overall impact of the neighbouring nodes, which is done using an additive attention module~\cite{bahdanau2014neural}. Finally, the output is concatenated with the node's history and road map data to produce a unified node representation vector $e_x$ fed to a decoder that constructs a predicted trajectory.

\subsection{Smooth Attention}

The \emph{smooth attention} approach~\cite{cao2022leveraging} presents a novel perspective on attention modules. Unlike traditional methods, it applies attention at each time step, following~\cite{vemula2018social}. Emulating human attention during deliberate tasks incorporates a smoothness constraint based on the hypothesis that attention does not frequently change over time. Previous research \cite{wolfe2000attention} shows that deliberate attention movements are slower due to internal limitations. This implies that attention does not frequently fluctuate during driving, as it falls under intentional movement. By incorporating the smoothness constraint, the \emph{smooth attention} approach enhances the attention mechanism, improving the selection of important information while disregarding less relevant input variables and aligning better with the characteristics of human attention.

\section{Smooth-Trajectron++} 

\begin{figure}[!h]
  \centering
  \includegraphics[scale=0.33]{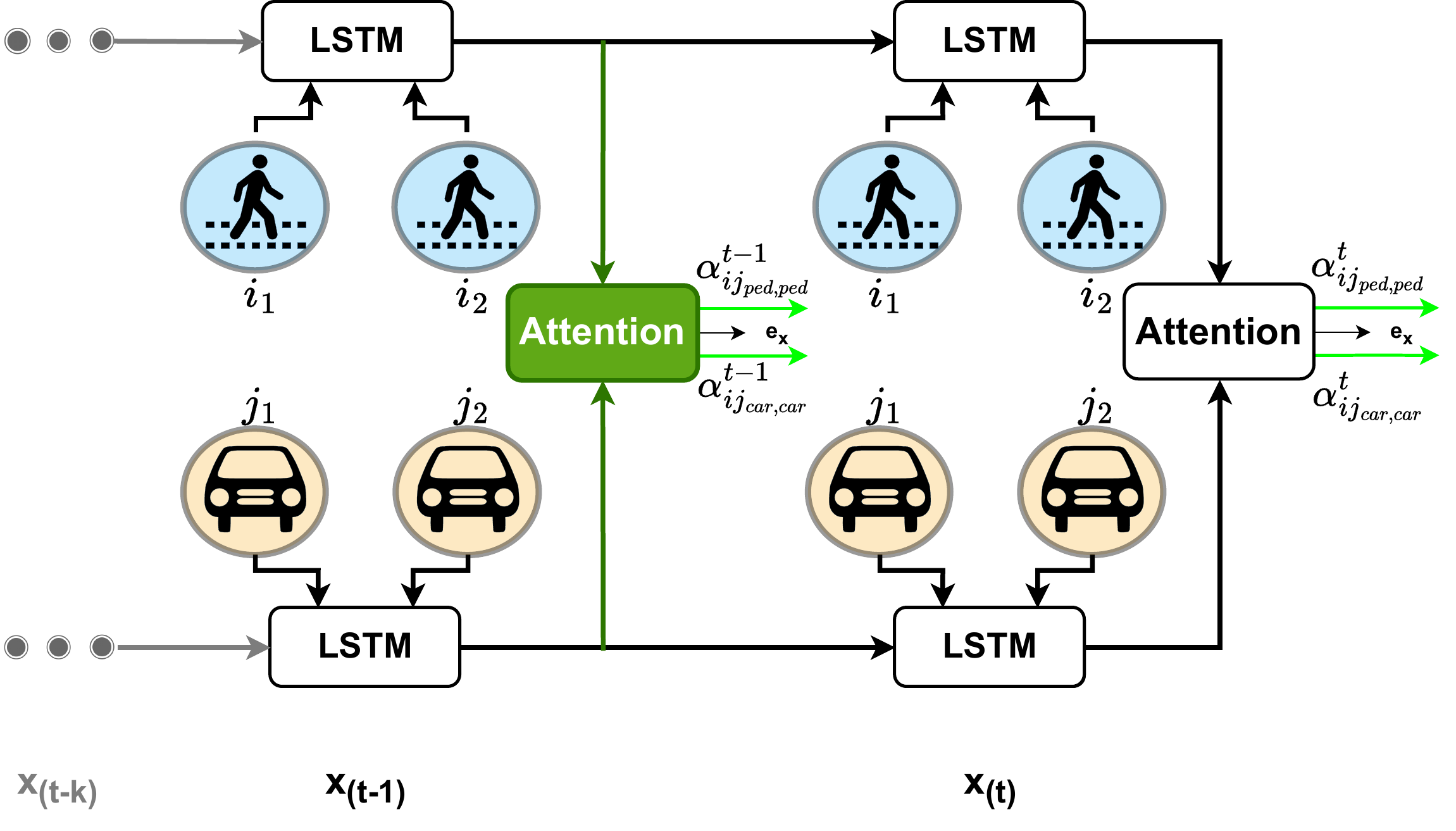}
  \caption{Edge Influence Encoder including Smooth-Attention (in green)}
  \label{fig:wholefig-encoder}
\end{figure}

In this section, we propose a way to apply the smooth attention module~\cite{cao2022leveraging} specifically to the \textit{Trajectron++} model. To do this, we alter the Edge Influence Encoder module of  \textit{Trajectron++}~\autoref{fig:encoderT++}, as this is the part where the social interactions are modelled, and the attention is applied.\\

Our approach\footnote{The source code is available at \hyperlink{https://github.com/fwesterhout/Smooth-Trajectron}{GitHub/Smooth-Trajectron++}}, which we call \emph{Smooth-Trajectron++}, is illustrated in \autoref{fig:wholefig-encoder}. At a high level, the original Edge Influence Encoder is expanded by applying the attention module at each time step in a similar fashion as in the \emph{smooth attention} model (the green highlighted box in \autoref{fig:wholefig-encoder}), where the outputs $\alpha^{\tau}_{i,j_{ab}}$ are the attention weights that are used to rank the importance the human agent $i$ assigns to the semantic class $j_{ab}$ for neighbouring agents of types $a$ and $b$ ($a$ and $b$ can stand for agent types such as cars or pedestrians) at the time $\tau$. All these attention weights from every time step are then used as an input for the added smoothing term in the loss function to incorporate the regularising of the attention by imposing a vectorial total variation penalty: 

\begin{equation}
\label{equation1}
    \mathcal{L}_{\mathrm{smooth}}(\alpha)=\sum_{\tau=t-T + 1}^{t} \sum_{i=1}^N \sqrt{\sum_j\left(\alpha_{i j}^t-\alpha_{i j}^{t-1}\right)^2}.
\end{equation}

To ensure that the attention weights are utilised during the model training process, we incorporate  $\mathcal{L}_{\mathrm{smooth}}$ into the original loss function $\mathcal{L}_0$~\cite{salzmann_trajectron_2020}:\\
   \begin{equation}
        \mathcal{L}_{new}= \mathcal{L}_{0} + \beta \mathcal{L}_{smooth}.
    \end{equation}
The scaling factor $\beta$ is introduced to fine-tune the influence of $\mathcal{L}_{smooth}$. By adjusting $\beta$, the model can be trained to effectively balance the contribution of the attention weights with the original loss function. 

The extra loss term $\mathcal{L}_{smooth}$ and associated additional calls to the attention module increase the number of computations and therefore have an effect on the training time, which is approximately 1.5 times slower compared to the original version of \emph{Trajectron++}.

\section{Results }

Our method is evaluated on two publicly available datasets: \emph{nuScenes} \cite{caesar_nuscenes_2020} and \emph{highD} \cite{krajewski2018highd}. In both scenarios, we trained and assessed both the original \emph{Trajectron++} model (which is a special case of \textit{Smooth-Trajectron++} for $\beta=0$) 
and the expanded model, with multiple versions of the latter, differentiated by five $\beta$-values ranging from $0.01$ to $10$. We also compared the obtained results with those originally reported for \textit{T++}~\cite{salzmann_trajectron_2020}, as these turned out to be substantially different from the results we obtained after directly reproducing \textit{T++} using the available source code and the same hyperparameters mentioned in the original article for the model’s training.

The experiments were performed on DelftBlue, the high-performance cluster of Delft University of Technology.

\subsection{nuScenes dataset}
The \emph{nuScenes} dataset~\cite{caesar_nuscenes_2020} consists of 1000 driving scenes in Boston and Singapore, characterised by their high traffic volumes and challenging driving situations. The driving scenes span 20 seconds each and are annotated at \SI{2}{\hertz}.\\

For this dataset, we evaluated the models according to three metrics: Final Displacement Error (FDE), Average Displacement Error (ADE) and Kernel Density Estimation of Negative Log Likelihood (KDE-NLL). These metrics are chosen as they were used in the original paper~\cite{salzmann_trajectron_2020}. First, FDE indicates how far off the model's predicted location is from the actual location at the end of a predicted trajectory. Second, ADE is particularly useful for evaluating the overall accuracy of a model's trajectory predictions, as it considers the entire predicted trajectory rather than just the final location. Finally, KDE-NLL is a valuable metric for evaluating the uncertainty of a model's predictions, as it measures how well the model can capture the true distribution of the data. Following~\cite{salzmann_trajectron_2020}, we calculate the three above metrics at prediction horizons of 1, 2, 3, and 4s. The FDE and the ADE outputs comprise the most likely single trajectory prediction, using the "Most Likely" output configuration as in \cite{salzmann_trajectron_2020}. 

There are two main agent classes in \emph{nuScenes}, pedestrians and vehicles. As their behaviour is significantly different, we evaluate the models on these classes separately.

\subsubsection{Pedestrian-only predictions}
Leftmost numbers in each column of \autoref{tab:FDE-nuScenes}-\autoref{tab:KDE-nuScenes} show the results for the predicted pedestrian trajectories. The numbers in bold represent the lowest metric values per prediction horizon, compared to the reproduced T++ results (the "T++ (rep)" row), which serve as a reference for comparative analysis. 

First, we found a significant gap between the reproduced T++ performance and the results reported and \textit{T++} paper. The FDE and ADE exhibit notable differences, especially at shorter prediction horizons. The reproduced KDE-NLL values also diverge significantly from the reported values. Several factors may contribute to this deviation; for example, a different version of \emph{nuScenes}, a discrepancy in used and reported hyper-parameters and model settings, or possible deviations introduced during the reproduction process, such as data downloading or package installation. Future research should more closely examine reproducibility of the original results and clarify potential causes of mismatches with the original findings. 

Second, the \textit{smooth attention} extensions of the reproduced \textit{T++} ($\beta=0.01$ to $\beta=10$) consistently outperform the baseline reproduced version of \textit{T++}. Tuning the scaling factor $\beta$ influences the error. Regarding the FDE, the parameter $\beta=0.1$ has the lowest error in all cases, except for the shared lowest error at the first prediction horizon (@1s) of $\beta=1.0$. The higher the $\beta$-value, the more it resembles the "T++ (rep)" reference values. However, in \autoref{tab:KDE-nuScenes}, the opposite seems to be happening; the "T++ (rep)" row shows the lowest values for almost all cases. An exception is the smooth version with $\beta=0.01$ @4s, where a marginal performance increase is seen. However, in general, in this pedestrian-only case, smooth attention does not improve this metric, although the decline for the smooth versions is minimal. The smoothing term might decrease the variety of predicted trajectory distributions, affecting the average and making it less similar to the ground truth. Further research is needed to explore this hypothesis in other pedestrian-only scenarios.

\begin{table}[h]
\centering
\caption{Results \emph{nuScenes} T++ pedestrian-only/vehicle-only: FDE (m)}
\label{tab:FDE-nuScenes}
\resizebox{\columnwidth}{!}{%
\begin{tabular}{@{}
>{\columncolor[HTML]{FFFFFF}}l 
>{\columncolor[HTML]{FFFFFF}}l 
>{\columncolor[HTML]{FFFFFF}}l 
>{\columncolor[HTML]{FFFFFF}}l 
>{\columncolor[HTML]{FFFFFF}}l @{}}
\toprule
{\color[HTML]{000000} \textit{Model}} & {\color[HTML]{000000} @1s}  & {\color[HTML]{000000} @2s}  & {\color[HTML]{000000} @3s}  & {\color[HTML]{000000} @4s}  \\ \midrule
{\color[HTML]{333333} T++~\cite{salzmann_trajectron_2020}}              & {\color[HTML]{333333} 0.014/0.07} & {\color[HTML]{333333} 0.17/0.45} & {\color[HTML]{333333} 0.37/1.14} & {\color[HTML]{333333} 0.62/2.20} \\ \midrule
{\color[HTML]{333333} T++ (rep.)}         & {\color[HTML]{333333} 0.168/0.430}  & {\color[HTML]{333333} 0.369/1.168} & {\color[HTML]{333333} 0.608/2.323} & {\color[HTML]{333333} 0.886/3.868} \\
{\color[HTML]{333333} $\beta=0.01$} & {\color[HTML]{333333} 0.157/\textbf{0.413}}  & {\color[HTML]{333333} 0.353/1.102} & {\color[HTML]{333333} 0.586/2.141} & {\color[HTML]{333333} 0.855/3.546} \\
{\color[HTML]{333333} $\beta=0.1$}  & {\color[HTML]{333333} \textbf{0.155}/0.419}  & {\color[HTML]{333333} \textbf{0.350/1.081}} & {\color[HTML]{333333} \textbf{0.580/2.122}} & {\color[HTML]{333333} \textbf{0.842/3.496}} \\
{\color[HTML]{333333} $\beta=0.5$}  & {\color[HTML]{333333} 0.159/0.421}  & {\color[HTML]{333333} 0.354/1.123} & {\color[HTML]{333333} 0.588/2.181} & {\color[HTML]{333333} 0.857/3.560} \\
{\color[HTML]{333333} $\beta=1.0$}  & {\color[HTML]{333333} \textbf{0.155}/0.448}  & {\color[HTML]{333333} 0.351/1.128} & {\color[HTML]{333333} 0.582/2.165} & {\color[HTML]{333333} 0.845/3.507} \\
{\color[HTML]{333333} $\beta=10$}   & {\color[HTML]{333333} 0.160/0.425}  & {\color[HTML]{333333} 0.366/1.149} & {\color[HTML]{333333} 0.607/2.190} & {\color[HTML]{333333} 0.876/3.539} \\ \bottomrule
\end{tabular}%
}
\end{table}

\begin{table}[h]
\centering
\caption{Results \emph{nuScenes} T++ pedestrian-only/vehicle-only: ADE (m)}
\label{tab:ADE-nuScenes}
\resizebox{\columnwidth}{!}{%
\begin{tabular}{@{}
>{\columncolor[HTML]{FFFFFF}}l 
>{\columncolor[HTML]{FFFFFF}}l 
>{\columncolor[HTML]{FFFFFF}}l 
>{\columncolor[HTML]{FFFFFF}}l 
>{\columncolor[HTML]{FFFFFF}}l @{}}
\toprule
{\color[HTML]{000000} \textit{Model}} & {\color[HTML]{000000} @1s}  & {\color[HTML]{000000} @2s}  & {\color[HTML]{000000} @3s}  & {\color[HTML]{000000} @4s}  \\ \midrule
{\color[HTML]{333333} T++~\cite{salzmann_trajectron_2020}}           & {\color[HTML]{333333} 0.021/-} & {\color[HTML]{333333} 0.073/-} & {\color[HTML]{333333} 0.15/-} & {\color[HTML]{333333} 0.25/-} \\ \midrule
{\color[HTML]{333333} T++ (rep)}      & {\color[HTML]{333333} 0.126/0.307}  & {\color[HTML]{333333} 0.221/0.632}  & {\color[HTML]{333333} 0.329/1.092} & {\color[HTML]{333333} 0.450/1.689} \\
{\color[HTML]{333333} $\beta=0.01$} & {\color[HTML]{333333} 0.116/\textbf{0.296}}  & {\color[HTML]{333333} 0.208/0.602}  & {\color[HTML]{333333} 0.314/1.021} & {\color[HTML]{333333} 0.432/1.559} \\
{\color[HTML]{333333} $\beta=0.1$}  & {\color[HTML]{333333} \textbf{0.114/0.302}}  & {\color[HTML]{333333} \textbf{0.206/0.597}}  & {\color[HTML]{333333} \textbf{0.311/1.012}} & {\color[HTML]{333333} \textbf{0.427/0.543}} \\
{\color[HTML]{333333} $\beta=0.5$}  & {\color[HTML]{333333} 0.118/0.301}  & {\color[HTML]{333333} 0.210/0.613} & {\color[HTML]{333333} 0.316/1.041} & {\color[HTML]{333333} 0.434/1.580} \\
{\color[HTML]{333333} $\beta=1.0$}  & {\color[HTML]{333333} \textbf{0.114}/0.319}  & {\color[HTML]{333333} 0.207/0.630}  & {\color[HTML]{333333} 0.312/1.048} & {\color[HTML]{333333} 0.428/1.575} \\
{\color[HTML]{333333} $\beta=10$}   & {\color[HTML]{333333} 0.118/0.303}  & {\color[HTML]{333333} 0.215/0.628}  & {\color[HTML]{333333} 0.325/1.055} & {\color[HTML]{333333} 0.445/1.586} \\ \bottomrule
\end{tabular}%
}
\end{table}

\begin{table}[h!]
\centering
\caption{Results \emph{nuScenes} T++ pedestrian-only/vehicle-only: KDE NLL}
\label{tab:KDE-nuScenes}
\resizebox{\columnwidth}{!}{%
\begin{tabular}{@{}
>{\columncolor[HTML]{FFFFFF}}l 
>{\columncolor[HTML]{FFFFFF}}l 
>{\columncolor[HTML]{FFFFFF}}l 
>{\columncolor[HTML]{FFFFFF}}l 
>{\columncolor[HTML]{FFFFFF}}l @{}}
\toprule
{\color[HTML]{000000} \textit{Model}} & {\color[HTML]{000000} @1s}  & {\color[HTML]{000000} @2s}  & {\color[HTML]{000000} @3s}  & {\color[HTML]{000000} @4s}  \\ \midrule
{\color[HTML]{333333} T++~\cite{salzmann_trajectron_2020}}           & {\color[HTML]{333333} -5.58/-4.17} & {\color[HTML]{333333} -3.96/-2.74} & {\color[HTML]{333333} -2.77/-1.62} & {\color[HTML]{333333} -1.89/-0.71} \\ \midrule
{\color[HTML]{333333} T++ (rep)}      & {\color[HTML]{333333} \textbf{-2.575}/-1.760} & {\color[HTML]{333333} \textbf{-1.530}/-0.604} & {\color[HTML]{333333} \textbf{-0.797}/0.235} & {\color[HTML]{333333} -0.230/0.927} \\
{\color[HTML]{333333} $\beta=0.01$} & {\color[HTML]{333333} -2.560/-1.861} & {\color[HTML]{333333} -1.519/\textbf{-0.726}} & {\color[HTML]{333333} -0.795/\textbf{0.108}} & {\color[HTML]{333333} \textbf{-0.240/\textbf{0.801}}} \\
{\color[HTML]{333333} $\beta=0.1$}  & {\color[HTML]{333333} -2.541/-1.856} & {\color[HTML]{333333} -1.502/-0.679} & {\color[HTML]{333333} -0.776/0.176} & {\color[HTML]{333333} -0.216/0.875} \\
{\color[HTML]{333333} $\beta=0.5$}  & {\color[HTML]{333333} -2.542/\textbf{-1.885}} & {\color[HTML]{333333} -1.505/-0.690} & {\color[HTML]{333333} -0.779/0.150} & {\color[HTML]{333333} -0.211/0.818} \\
{\color[HTML]{333333} $\beta=1.0$}  & {\color[HTML]{333333} -2.549/-1.880} & {\color[HTML]{333333} -1.507/-0.679} & {\color[HTML]{333333} -0.785/0.173} & {\color[HTML]{333333} -0.226/0.868} \\
{\color[HTML]{333333} $\beta=10$}   & {\color[HTML]{333333} -2.480/-1.861} & {\color[HTML]{333333} -1.471/-0.661} & {\color[HTML]{333333} -0.759/0.173} & {\color[HTML]{333333} -0.205/0.845} \\ \bottomrule
\end{tabular}%
}
\end{table}

\subsubsection{Vehicle-only predictions}
Rightmost numbers in each column of \autoref{tab:FDE-nuScenes}-\autoref{tab:KDE-nuScenes} show the results for the predicted vehicle trajectories. Similarly to the previous pedestrian-only case, a general FDE and ADE decline is seen along the $\beta$-versions of the \textit{Smooth-Trajectron++}. The ADE values of the \textit{T++} paper are missing in \autoref{tab:ADE-nuScenes}, as they are not reported by the authors in the original article. The version with $\beta=0.01$ holds the lowest value for the prediction horizon of 1 second, while $\beta=0.1$ has a minor error for the remaining prediction horizons. Contrary to the pedestrian-only predictions in \autoref{tab:KDE-nuScenes}, \textit{Smooth-Trajectron++} on the vehicle-only forecasts has better KDE-NLL numbers than the reproduced model, which indicates that the model is better able to match the original distribution of predicted trajectories with the inclusion of the smooth-attention term in the loss function. Furthermore, this can be said for all $\beta$-factors, while in this case, the $\beta=0.01$ has the lowest values.

\subsection{highD dataset}
To evaluate the models on the \emph{highD} dataset, we used the previously proposed benchmarking framework~\cite{benchmarking}.
This framework was designed to benchmark prediction models in \textit{gap acceptance} scenarios, i.e. situations where drivers decide whether to enter a gap in traffic or wait for the next opportunity, such as when a car approaches an intersection and decides whether to turn left immediately or wait for a break in oncoming traffic. In case of highD dataset, we investigated the predictions of gap acceptance in lane-change decisions using a restricted version of highD (see~\cite{benchmarking} for details).


The framework~\cite{benchmarking} allowed us to use two methods of splitting the highD data into training and testing sets: the random split and the critical split. The first method randomly splits the data for testing and training. In contrast, the second method deliberately selects the most unusual behaviour for testing, such as accepting a very small gap or rejecting a large gap. This latter testing scenario, therefore, tests the model's ability to extrapolate to situations that lie outside its training distribution, which is generally considered to be a more difficult task~\cite{barnard1992extrapolation}. Also, small accepted gaps can be regarded as safety-critical scenarios, which is especially important when developing safe and reliable prediction models.

Furthermore, the framework allowed us to test the models with varying number of input time steps($n_I$) to study the input-dependability of the tested models; we used $n_I = 2$ and $n_I = 10$.

In addition to the metrics used for the \emph{nuScenes} dataset, the gap acceptance benchmark includes an additional metric, the Area under the Receiver-Operator Curve (AUC), used to evaluate the performance of binary classification models (here between accepted and rejected gaps).

\begin{figure}[]
    \centering
    \input{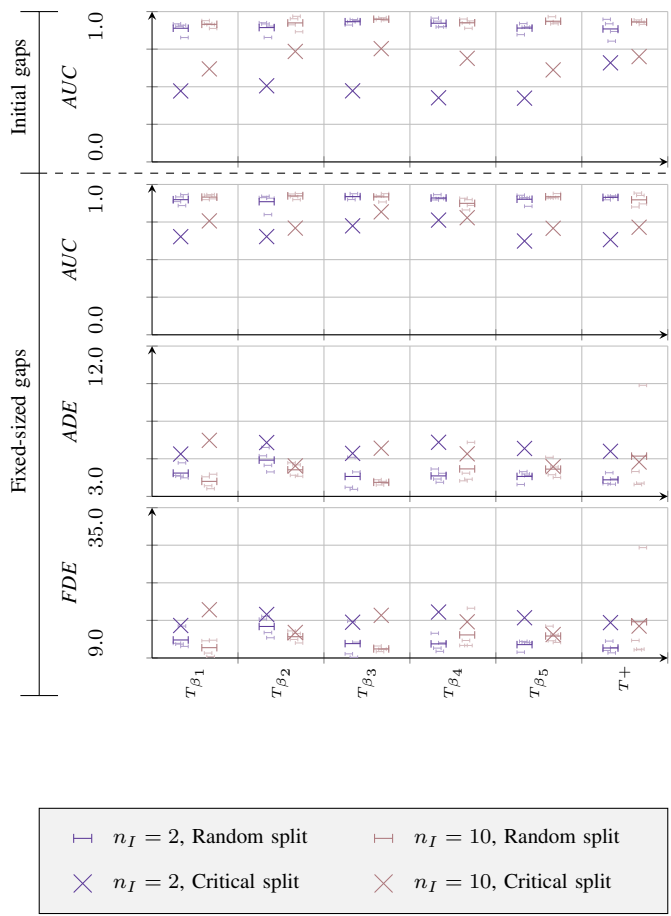}
    \caption{The performance of \textit{Smooth-Trajectron++} and \emph{T++} on the \emph{highD} dataset: $\beta_{1-5}$ refer to the $\beta $-values of 0.01, 0.1, 0.5, 1.0 and 10 respectively}
    \label{fig:benchmarkfig}
\end{figure}

 First, we analyse the performance of the models at predicting lane-change decisions at initial gaps, i.e. at the start of the interaction. 
Here, only in the case of $\beta = 0.5$ there is a notable increase in AUC for the random split compared to $T+$ in both $n_I$-instances. For the critical split, almost all AUC-values are lower than $T+$, except for $\beta = 0.1$ and $\beta = 0.5$ at $n_I = 10$ where it is slightly higher. Generally, the $\beta$-term does not seem to increase the performance of the base model.

Second, we investigated models' predictions of lane changes in \emph{highD} at the fixed-sized gaps, as defined in \cite{benchmarking}. Looking at the random splits, again for $\beta = 0.5$ there is an increase in AUC for both $n_I$-situations. The changes for the other $\beta$-versions are not consistently different when compared to $T+$, having minor fluctuations to perform slightly better or slightly worse. Concerning the critical split, all $\beta$-versions but $\beta = 10$ perform better than the base model, where $\beta = 10$ performs very similarly to $T+$. Also, the difference between  $n_I = 2$ and $n_I = 10$ is logical, as the latter consistently has a higher AUC than the former. 

For both the FDE and ADE, all the $\beta$-versions are outperforming the $T+$-model at $n_I = 10$ on the random split. However, this seems due to one extremely high value of one of the random splits of $T+$ (each random split consists of three sub-splits, which are averaged to minimize the effect of randomness). This could be an outlier, caused by an error in the training process. At $n_I = 2$, the $\beta$-values under-perform compared to $T+$ for the random split, indicating no significant improvement. At the critical split, only at $\beta = 0.1$ and $\beta = 10$ both ADE and FDE values are lower at $n_I = 10$. In general, there is no clear improvement regarding these metrics across the various $\beta$-values.\\

Overall, in \emph{highD} lane-change prediction experiments, there are instances of both better and worse performance of \textit{Smooth-Trajectron++} compared to T++, indicating no consistent benefits of adding smooth attention to T++. This is in contrast to the nuScenes results, which may stem from fundamental differences in the datasets. While the \emph{nuScenes} dataset encompasses a wide range of data with cars and pedestrians, \emph{highD} mainly focuses on cars. The application of the smoothing term $\beta$ in the \textit{Smooth-Trajectron++} model relies on the attention module that compares different semantic classes of traffic participants. In datasets where one class dominates, the smoothing term may not yield tangible improvements.

\section{Conclusion}



This paper proposed \textit{Smooth-Trajectron++}, a trajectory prediction model based on an existing state-of-the-art model \textit{Trajectron++}~\cite{salzmann_trajectron_2020} in which we incorporated a cognitively-inspired smooth attention module~\cite{cao2022leveraging}. We demonstrated that our smooth-attention version of T++ can achieve increased performance on the large-scale dataset nuScenes, but does not result in tangible improvements on the highD dataset. This suggests that the smooth attention approach seems to be more suitable for large-scale multi-agent datasets with multiple agent types rather than on datasets with few traffic agents of mostly the same type. 
Hence, the concept of \emph{smooth attention} might be better applied to models where the attention module is implemented over individual agents and not semantic classes. Nevertheless, our results further strengthen previous work \cite{schumann_using_2023,song_research_2022,cao2022leveraging}, indicating that including cognitive insights can allow better predictions of human behavior in traffic.

\bibliographystyle{jabbrv_ieeetr}
\bibliography{IEEEabrv,main.bbl}

\end{document}